# R-WhONet: Recalibrated Wheel Odometry Neural Network for Vehicular Positioning using Transfer Learning

Uche Onyekpe[1,2*], Alicja Szkolnik[3], Vasile Palade[2], Stratis Kanarachos[4], Michael E. Fitzpatrick[4]

[1] *Department of Computer and Data Science, School of Science, Technology and Health, York St John University, York YO31 7EX, UK*
[2] *Research Centre for Data Science, Coventry University, Priory Road, Coventry, CV1 5FB, UK*
[3] *University of Birmingham, B15 2TT, UK*
[4] *Faculty of Engineering, Coventry University, Priory Road, Coventry, CV1 5FB, UK*
Email: u.onyekpe@yorksj.ac.uk, ais103@student.bham.ac.uk, ab5839@coventry.ac.uk, ab8522@coventry.ac.uk, ab6856@coventry.ac.uk

## Abstract

This paper proposes a transfer learning approach to recalibrate our previously developed Wheel Odometry Neural Network (WhONet) for vehicle positioning in environments where Global Navigation Satellite Systems (GNSS) are unavailable. The WhONet has been shown to possess the capability to learn the uncertainties in the wheel speed measurements needed for correction and accurate positioning of vehicles. These uncertainties may be manifested as tyre pressure changes from driving on muddy and uneven terrains or wheel slips. However, a common cause for concern for data-driven approaches, such as the WhONet model, is usually the inability to generalise the models to a new vehicle. In scenarios where machine learning models are trained in a specific domain but deployed in another domain, the model's performance degrades. In real-life scenarios, several factors are influential to this degradation, from changes to the dynamics of the vehicle to new pattern distributions of the sensor's noise, and bias will make the test sensor data vary from training data. Therefore, the challenge is to explore techniques that allow the trained machine learning models to spontaneously adjust to new vehicle domains. As such, we propose the **R**ecalibrated-**Wh**eel **O**dometry neural **Net**work (**R-WhONet**), based on transfer learning, that adapts the WhONet model from its source domain (a vehicle and environment on which the model is initially trained) to the target domain (a new vehicle on which the trained model is to be deployed). Through a performance evaluation on several GNSS outage scenarios – short-term complex driving scenarios such as on roundabouts, sharp cornering, hard-brake and wet roads (drifts), and on longer-term GNSS outage scenarios of 30s, 60s, 120s and 180s duration – we demonstrate that a model trained in the source domain does not generalise well to a new vehicle in the target domain. However, we show that our new proposed framework improves the generalisation of the WhONet model to new vehicles in the target domains by up to 32%.

## 1. Introduction

Autonomous and automated vehicles and robots alike require accurate positioning and orientation information for effective and safe navigation [1]–[3]. Various approaches have adopted different sensor combinations to aid the tracking of these vehicles in areas deprived of Global Navigation Satellite Systems (GNSS) signals [4]–[7]. Examples of such environments are tunnels, bridges, dense tree canopies, and tall buildings which block the direct line of sight of the GNSS transmitter and receiver during signal transmission [3]. Sensor combinations for low-cost Inertial Navigation System (INS) sensors generally consist of an accelerometer and a gyroscope. The former measures the acceleration of the vehicle, and the latter measures the vehicle's attitude rate. INS measurements can, however, be skewed by noises such as thermo-mechanical, flicker noises, calibration errors, etc., which are amplified when integrating the vehicle's acceleration to displacement, and orientation rate to orientation [8]. Notably, this error exponentially accumulates over time during position Dead Reckoning (DR), to provide a poor navigation output [8]. Hence, continuous updates are required for position error correction.

Several machine-learning-based techniques have been proposed to learn the error drift over time to provide an accurate estimation of the vehicle's position and orientation. Such proposed techniques include Recurrent-Neural-Networks-(RNN) based models in [3], [8]–[12], MFNN-based models in [13]–[17], RBFNN-based models in [18], [19] and the IDNN in [20]. Despite considerable research into improving the performance of low-cost INS, the positioning issue remains a challenge in need of cost-effective solutions.

* Corresponding author Uche Onyekpe: u.onyekpe@yorksj.ac.uk

Autonomous vehicles are supported by sensors such as wheel encoders needed for advanced driving-assist systems. The wheel encoder that measures the speed of the vehicle's wheels is considered a simpler and more desirable approach for position estimation compared to low-cost accelerometers [4], as it requires fewer integration steps to establish a vehicle's position [4]. Nevertheless, wheel odometry also accumulates errors due to travel on uneven surfaces; wheel slippage caused by travel on wet and slippery roads, skidding, etc; and unequal wheel diameters caused by changes in the tyre pressures. Reference [21] showed that the errors present within the position estimation obtained from the wheel speed data can be learned by the Long-Short-Term Memory (LSTM) neural network even in complex driving environments such as roundabouts, successive left and right turns, wet roads, etc. In c, a Wheel Odometry Neural Network (WhONet) was proposed and shown to provide better estimations in both complex driving scenarios and longer-term GNSS outages of up to 180s, with an accuracy averaging 8.62m after 5.6km of travel.

However, traditional machine learning techniques have so far been designed to work in isolation [22], [23]. The WhONet has been trained to learn the positional errors associated with a particular car size, sensor error pattern, as well as limited worn tyre states and tyre pressure variations [11]. However, there is a limit to the accuracy of the positional uncertainty the WhONet model can learn, as the sizes in which vehicles come are numerous and the wearing of tyres and changes in tyre pressures are uncontrollable. An approach to addressing this problem could be to train a model to fit the dynamics of each vehicle type on the road. However, this method is too expensive and impractical, particularly due to the different tyre states of each vehicle at different times of use. This paradigm can, however, be overcome through the use of transfer learning to recalibrate a pre-trained model on a single vehicle to adapt to the dynamics and tyre state of the new vehicles on which it is to be used.

Transfer learning within the context of machine learning has mostly found motivations in complex problems requiring a large amount of data for accurate performance [24]. The difficulty in getting large amounts of labelled data and the high cost of labelling the data has made it challenging to sufficiently train accurate machine learning models for different domain applications [25]. Through the use of pre-trained models such as on the ImageNet [22], with millions of images relating to different categories, it has been shown that transfer learning can be used to transfer features and parameters of the pre-trained model to a new domain application [26]–[28]. Transfer learning transcends specific task and domain applications to leverage knowledge learnt from a pre-trained model to solve problems in different application domains [29]–[35]. It is with such inspiration that we propose in this work the Recalibrated Wheel Odometry neural Network (R-WhONet). This research builds on the research done in [11] by providing a framework called R-WhONet (Recalibrated Wheel Odometry neural Network) for the adaptation of the WhONet model to vehicles with different feature characteristics such as tyre pressures, worn-out tyre state, driving behaviour, size, etc. We study the ability of the R-WhONet model to accurately and robustly adapt the WhONet model to new vehicles and scenarios using transfer learning, by evaluating its performance on longer-term GNSS outage scenarios using the CUPAC dataset (as it is characterised by different vehicle dynamics).

R-WhONet leverages the knowledge transfer capabilities of transfer learning to recalibrate the WhONet model to adapt to different vehicle dynamics and tyre states. We show that R-WhONet is able to consistently adapt a generic pre-trained WhONet model (which we refer to as G-WhONet in the rest of the paper for simplicity) to provide better uncertainty estimations whilst providing accuracies similar to a WhONet model (referred to as S-WhONet in the rest of the paper for simplicity) trained specifically for the new vehicle.

The contributions of this paper are summarised below.

1. A transfer learning based approach is proposed to adapt the WhONet model to other vehicles, thus improving its generalisation to such vehicles. The proposed approach, called R-WhONet, enhances the performance of a G-WhONet model in predicting the positioning errors of autonomous vehicles using wheel encoders in GNSS-deprived environments.
2. The accuracy and robustness of the R-WhONet model are evaluated on several short-term challenging driving scenarios as well as longer-term GNSS-outage scenarios of up 180 s from another dataset (CUPAC vehicle) with different motion dynamics. The obtained results demonstrate that the R-WhONet is able to provide accuracies similar to the S-WhONet model.

The rest of the paper is structured as follows: In Section 2, we provide mathematical definitions of transfer learning and other associated terms. In Section 3, we introduce the vehicular localisation problem and describe all the models used in this research. Section 4 defines the dataset and metrics used in evaluating the performance of the models and provides more information on how the models are trained. Subsequently, Section 5 discusses the results obtained during the evaluation of G-WhONet, S-WhONet, R-WhONet, and the physics models; and Section 6 concludes the study described in the paper.

## 2. Proposed Methodology

## 2.1 Domain representations and Transfer Learning

We define a Domain $D$ as a two-element tuple made up of a feature space $\mathcal{X}$ (which is characterised by a specific set of tyre pressure, worn-out state, and vehicle dynamics), and a marginal probability $P(\boldsymbol{X})$ such that $\boldsymbol{X}$ is a learning sample. Mathematically, the domain, D can be defined as $D = \{\mathcal{X}, P(\boldsymbol{X})\}$

Where $\boldsymbol{X} = \{\mathcal{X}_1, \dots, \mathcal{X}_m\}, \mathcal{X}_t \epsilon \mathcal{X}$,

here, $\mathcal{X}_t$ refers to a specific vector describing each sample data point at each time $t$. The specific vector $X$ describes the wheel speed measurements of the four wheels of the vehicle.

Task $T$ is also defined as a two-element tuple with an objective estimation function $n$ and a label space $\mathcal{Y}$. The objective function can thus be expressed probabilistically as $P(\boldsymbol{Y}|\boldsymbol{X})$. Mathematically, the task, $T$ can be defined as $T = \{\mathcal{Y}, P(\boldsymbol{Y}|\boldsymbol{X})\} = \{\mathcal{Y}, n\}$.

Where $\boldsymbol{Y} = \{\boldsymbol{y}_1, \dots, \boldsymbol{y}_m\}, \boldsymbol{y}_t \epsilon \mathcal{Y}$,

here, $\boldsymbol{y}_t$ refers to a specific vector describing the label of a sample data point at each time $t$. The specific vector $\boldsymbol{y}$ describes the position uncertainty measurement of the vehicle, and $\boldsymbol{Y}$ represents the corresponding labels of $\boldsymbol{X}$.

We also define the following necessary representations:

**Source Domain $\boldsymbol{D_S}$** – In this research, the source domain is defined as the vehicle whose dynamics and states are used to train the WhONet model.

**Source task $\boldsymbol{T_S}$** – The source task is defined as the uncertainty measurement of the vehicle's position due to the dynamics and states of the source vehicle.

**Target Domain $\boldsymbol{D_T}$** – We define the task domain as a new vehicle whose dynamics and states are different from the source domain.

**Target Task $\boldsymbol{T_T}$** – The target task is defined as the uncertainty measurement of the vehicle's position due to the dynamics and states of the new vehicle.

Given a source task $T_S$, a reciprocal source domain $D_S$, a target task $T_T$, and a target domain $D_T$, the aim of transfer learning is to enable the discovery of the target probability distribution $P(\boldsymbol{Y}_T|\boldsymbol{X}_T)$ in $D_T$ with the statistics obtained from $D_S$ and $T_S$ where $D_S \neq D_T$ or $T_S \neq T_T$. $\boldsymbol{Y}_T$ refers to the position uncertainty estimation and $\boldsymbol{X}_T$ refers to the wheel speed measurements of the four wheels.

The following conditions need to be satisfied for transfer learning:

1. $\mathcal{X}_s \neq \mathcal{X}_T$, the feature spaces of the target and source domain are not the same.
2. $P(\boldsymbol{X}_s) \neq P(\boldsymbol{X}_T)$, the marginal probability distributions of the target and source domain are different.
3. $\mathcal{Y}_s \neq \mathcal{Y}_T$, the labels between the source and target tasks vary.
4. $P(\boldsymbol{Y}_s|\boldsymbol{X}_s) \neq P(\boldsymbol{Y}_T|\boldsymbol{X}_T)$, the conditional probability distributions of the target and source tasks differ.

## 2.2. Physics model

During the positional tracking of vehicles using dead reckoning, the measurements are usually provided in the sensor's frame and then transformed into the navigation frame for positional tracking using the transformation matrix expressed in Equation (1).

$$R^{nb} = \begin{bmatrix} cos\Psi & -sin\Psi & 0 \\ sin\Psi & cos\Psi & 0 \\ 0 & 0 & 1 \end{bmatrix} \tag{1}$$

where $\Psi$ represents the yaw of the vehicle. The roll and the pitch in this case are neglected as this study investigates vehicular navigation on the horizontal plane.

The wheel speed sensors measure the angular velocity of the vehicle's wheels at a given time $t$. However, the tyre's diameter is prone to uncertainties related to tyre wearing, changes in tyre pressure, and wheel slip, which affect the accuracy of the displacement estimation from the wheel speed measurements. The following equations describe the errors which can affect the vehicle speed calculations.

$$\hat{\omega}^b_{whrl} = \omega^b_{whrl} + \varepsilon^b_{whrl} \tag{2}$$

$$\hat{\omega}^b_{whrr} = \omega^b_{whrr} + \varepsilon^b_{whrr} \tag{3}$$

$$\hat{\omega}^b_{whfl} = \omega^b_{whfl} + \varepsilon^b_{whfl} \tag{4}$$

$$\hat{\omega}^b_{whfr} = \omega^b_{whfr} + \varepsilon^b_{whfr} \tag{5}$$

Where $\hat{\omega}^b_{whrl}$, $\hat{\omega}^b_{whrr}$, $\hat{\omega}^b_{whfl}$ and $\hat{\omega}^b_{whfr}$ are the initial noisy wheel speed measurements of the rear left, rear right, front left, and front right wheels, respectively. $\varepsilon^b_{whrl}$, $\varepsilon^b_{whrr}$, $\varepsilon^b_{whfl}$ and $\varepsilon^b_{whfr}$ are the corresponding uncertainties and $\omega^b_{whrl}$, $\omega^b_{whrr}$, $\omega^b_{whfl}$ and $\omega^b_{whfr}$ are the respective wheel speed measurements excluding the errors.

Equations (6) and (7) show the calculation of the wheel speed of the rear axle.

$$\widehat{\omega}^{b}_{whr} = \frac{\omega^{b}_{whrr} + \omega^{b}_{whrl}}{2} + \frac{\varepsilon^{b}_{whrr} + \varepsilon^{b}_{whrl}}{2} \quad (6)$$

Assuming the expression $\frac{\varepsilon^{b}_{whrr}+\varepsilon^{b}_{whrl}}{2}$ as $\varepsilon^{b}_{whr}$ and $\frac{\omega^{b}_{whrr}+\omega^{b}_{whrl}}{2}$ as $\omega^{b}_{whr}$

$$\widehat{\omega}^{b}_{whr} = \omega^{b}_{whr} + \varepsilon^{b}_{whr} \quad (7)$$

The linear velocity of the vehicle can be calculated using the $v = \omega r$ formula, with $r$ being a constant which enables the mapping of the wheel speed of the rear axle to the linear velocity of the vehicle:

$$v^{b}_{wh} = \omega^{b}_{whr} r + \varepsilon^{b}_{whr} r \quad (8)$$

Take $\varepsilon^{b}_{whr} r \ as \ \varepsilon^{b}_{whr,v}$

$$v^{b}_{whr} = \omega^{b}_{whr} r + \varepsilon^{b}_{whr,v} \quad (9)$$

By integrating the vehicle's velocity obtained from $Equation$ 9 and incrementally updating for continuous tracking, it is possible to determine the displacement of the vehicle in the body frame. $\varepsilon^{b}_{whr,x}$ in $Equation$ 10 is the integral of $\varepsilon^{b}_{whr,v}$ from $Equation$ 9.

$$x^{b}_{whr} = \int_{t-1}^{t} (\omega^{b}_{whr} r) + \varepsilon^{b}_{whr,x} \quad (10)$$

During the presence of GNSS signals, the position uncertainty estimation $\varepsilon^{b}_{whr,x}$ can be found as shown in Equation (11). The task thus becomes that of estimating $\varepsilon^{b}_{whr,x}$ during GNSS outages needed to correct the vehicle's displacement $x^{b}_{whr}$.

$$\varepsilon^{b}_{whr,x} \approx x^{b}_{whr} - x^{b}_{GNSS} \quad (11)$$

The vehicle's true displacement, $x^{b}_{GNSS}$, is found using Vincenty's formula for geodesics on an ellipsoid [3], based on the vehicle's longitudinal and latitudinal positional information [3], [36], [37]. The accuracy of $x^{b}_{GNSS}$ is however limited to the accuracy of the GNSS used in this study which, according to [38], is defined as ±3m.

## 2.3. WhONet Model

Figure 1 shows the prediction block of the WhONet approach[1] during the absence of the GNSS signal. Here NED refers to the North-East-Down coordinate defining the navigation frame. WhONet is based on a simple RNN (see [11] for more details)

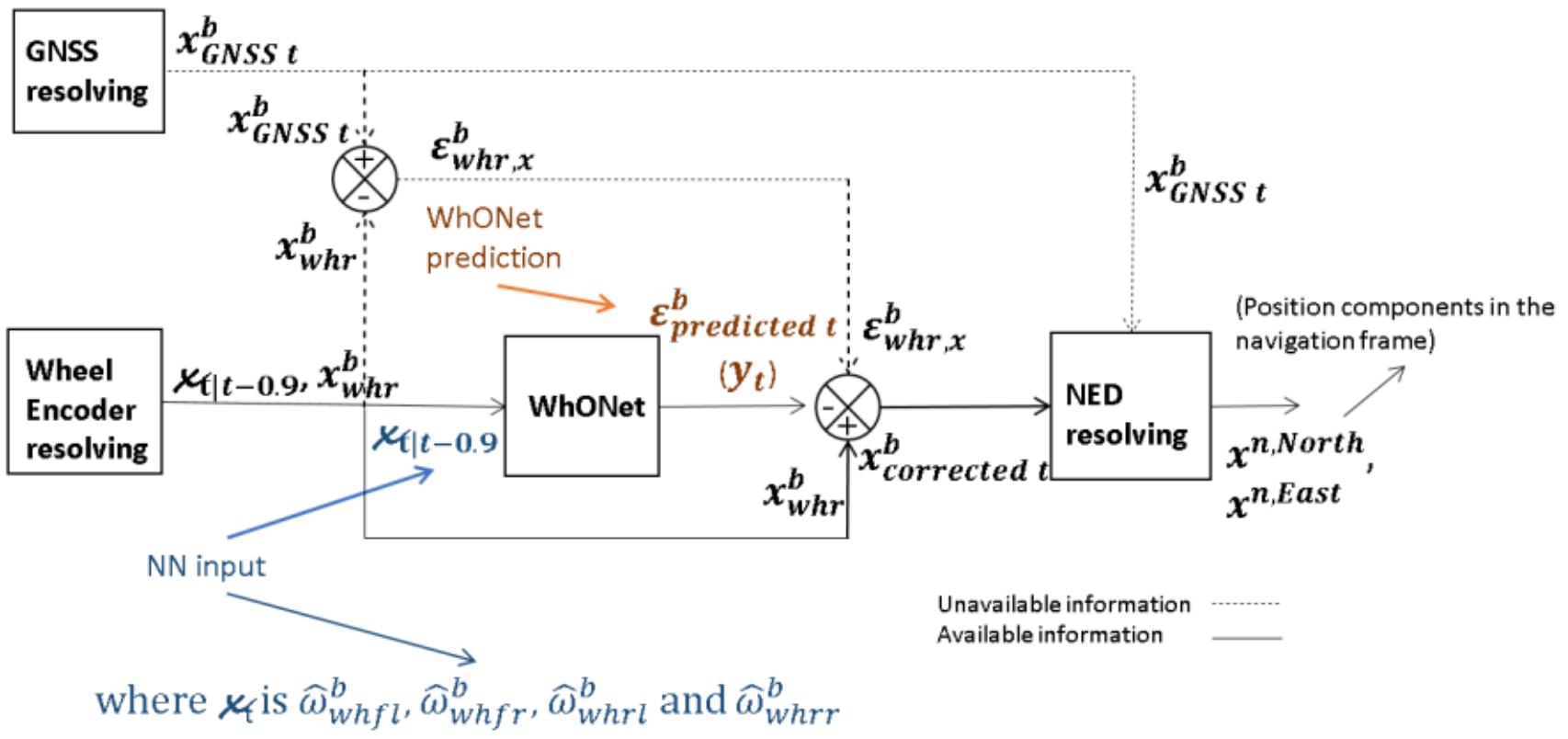

*Figure 1. WhONet's position-prediction block during GNSS outages.*

The simple RNN introduced by Rumelhart *et al*. in [39] has found use in many time-series applications [40]–[44], as it has the unique capability of discovering associations within sequential data. This simple RNN comprises feedback loops which connect relationships learnt in the past, creating what is commonly referred to as memory. The information acquired within the sequential dimension of the data is stored within the hidden state of the RNN and extended across the defined number of times steps before being mapped to the output. The architecture of a simple RNN is displayed in Figure 2 whilst the equations governing its operations are presented in $Equations$ (12) $and$ (13).

---

[1] WhONet's implementation can be found at https://github.com/onyekpeu/WhONet.

$$\boldsymbol{h}_t = tanh(\boldsymbol{U}_h \boldsymbol{h}_{t-1} + \boldsymbol{W}_x \boldsymbol{\mathcal{X}}_t + \boldsymbol{b}_h) \tag{12}$$

$$y_t = \sigma(\boldsymbol{W}_o \boldsymbol{h}_t + \boldsymbol{b}_o) \tag{13}$$

$\boldsymbol{\mathcal{X}}_t$ being defined as the input feature vector to the RNN, $\boldsymbol{y}$ as the output vector, $\sigma$ as the sigmoid activation (non-linearity) function, $\boldsymbol{h}_{t-1}$ as the previous state, $\boldsymbol{U}_h$ as the hidden weight matrix, $\boldsymbol{W}_x$ as the input weight matrix, $\boldsymbol{W}_o$ as the output weight matrix and $\boldsymbol{b}$ as the bias vector.

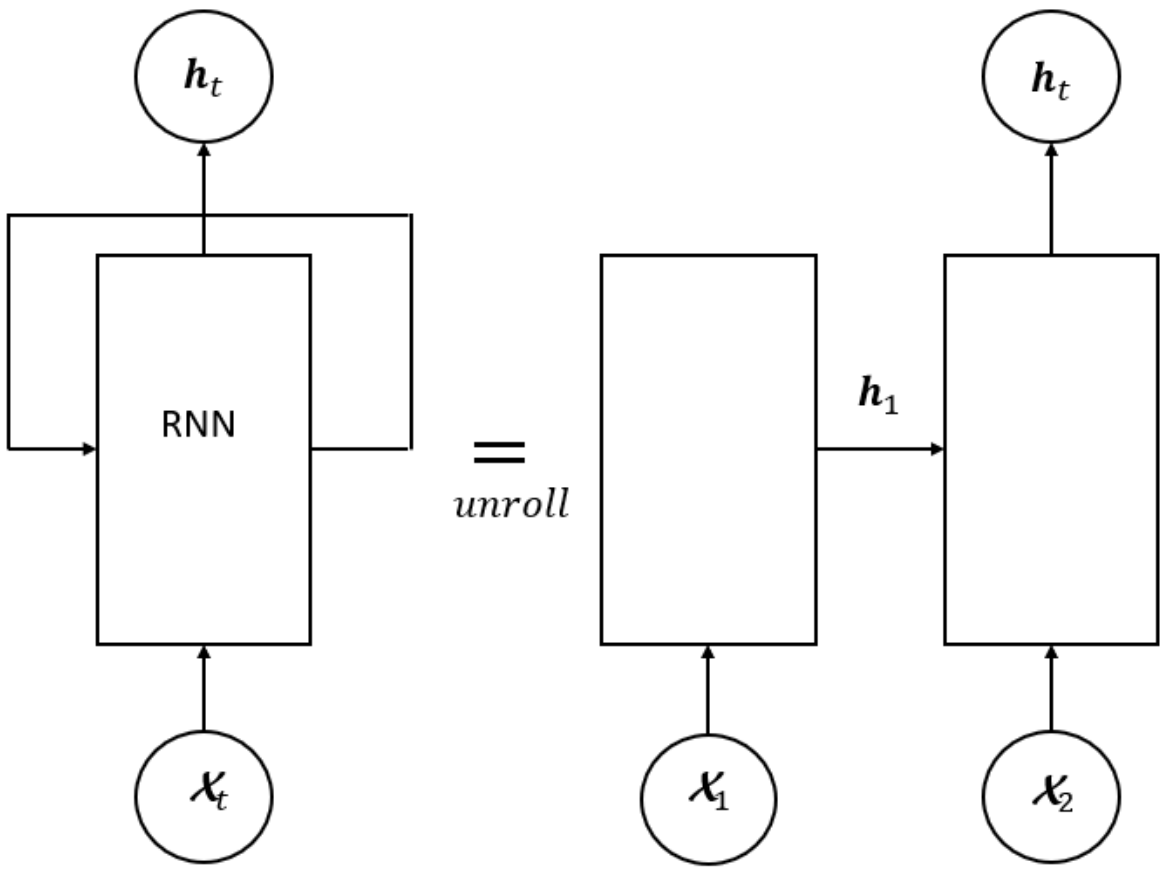

***Figure 2.*** *The unrolled architecture of the simple RNN used in WhONet.*

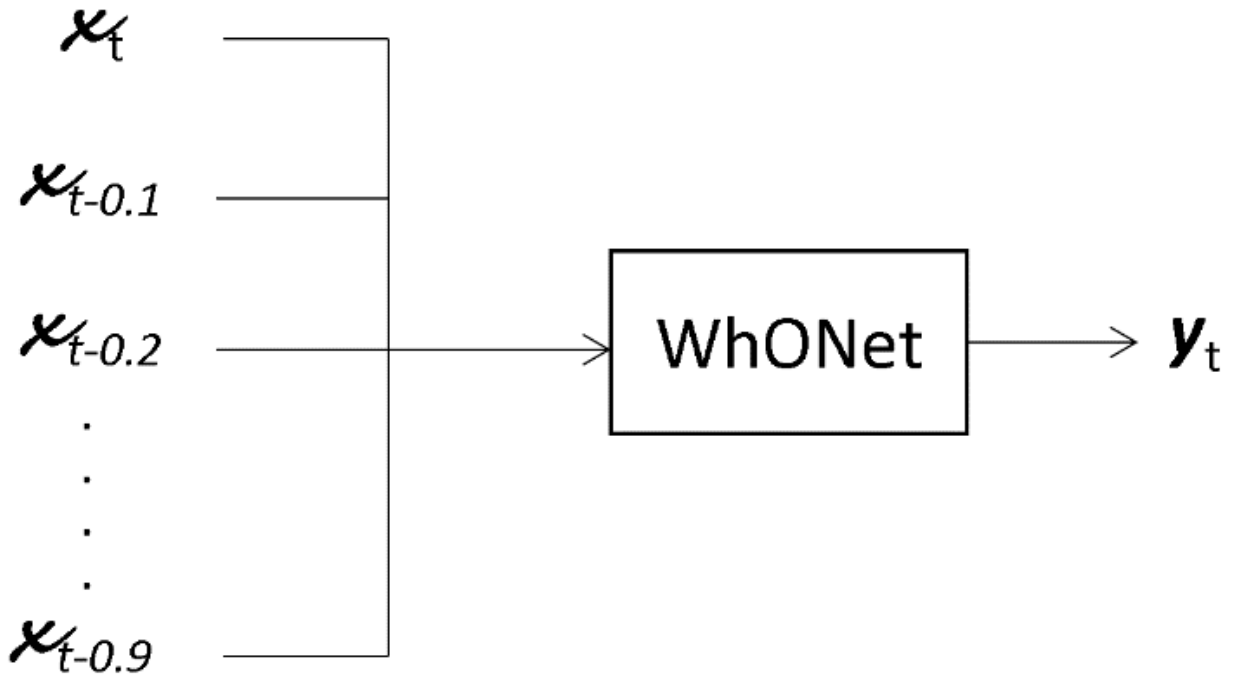

***Figure 3***. *WhONet's learning scheme*

At any time, $t$:

$$\mathcal{X}_1 = \boldsymbol{\mathcal{X}}_{t|t-0.9} \tag{14}$$

$$\mathcal{X}_2 = \boldsymbol{\mathcal{X}}_{t-1|t-1.9} \tag{15}$$

here $\mathcal{X}_t$ is further described as the temporal-sequenced input feature to the RNN at time $t$, which is characterised by the wheel speed information from the four wheels of the vehicles: $\widehat{\omega}^b_{whrl}$, $\widehat{\omega}^b_{whrr}$, $\widehat{\omega}^b_{whfl}$ and $\widehat{\omega}^b_{whfr}$, and $\mathcal{X}_1$ and $\mathcal{X}_2$ are the temporal-sequenced features at timestep $t$ and $t-1$ respectively as shown in Figure 2 and 3. The RNN is tasked with predicting $\boldsymbol{y}_t$, which is the error $\varepsilon^b_{whr,x}$ between the wheel-speed-derived displacement $x^b_{whr}$ and the GNSS-derived displacement $x^b_{GNSS}$.

## 2.4 S-WhONet (Specific - Wheel Odometry neural Network) Model

This refers to the WhONet model which is pre-trained only on the vehicle in which it is to be deployed *i.e.*, the target domain becomes the source domain. For every new vehicle, change in vehicle state, and/or dynamics, a new S-WhONet model is trained to more accurately predict the position estimation of the new vehicle at every time $\boldsymbol{t}$.

As such, S-WhONet is characterised by the following definitions:

1. $\mathcal{X}_s = \mathcal{X}_T$, the feature spaces of the target and source domain are the same.
2. $P(\boldsymbol{X}_S) = P(\boldsymbol{X}_T)$, the marginal probability distributions of the target and source domain are the same.
3. $\mathcal{Y}_s = \mathcal{Y}_T$, the labels between the source and target tasks are the same.
4. $P(\boldsymbol{Y}_T|\boldsymbol{X}_T) = 0$, the conditional probability distributions of the target domain are not found, rather, the conditional probability distributions of the source domain are used directly in the target domain.

## 2.5 G-WhONet (Generic Wheel Odometry neural Network) Model

This describes the WhONet model trained on the source vehicle, but that does not undergo any form of adaptation before deployment on the target vehicle.
G-WhONet is characterised by the following:
1. $\mathcal{X}_S \neq \mathcal{X}_T$, the feature spaces of the target and source domain are different.
2. $P(\boldsymbol{X}_S) \neq P(\boldsymbol{X}_T)$, the marginal probability distributions of the target and source domain are different.
3. $\mathcal{Y}_S \neq \mathcal{Y}_T$, the labels between the source and target tasks are not the same.
4. $P(\boldsymbol{Y}_T|\boldsymbol{X}_T) = 0$ the conditional probability distributions of the target are not found, rather, the conditional probability distribution of the source domain $P(\boldsymbol{Y}_S|\boldsymbol{X}_S)$ is used in the target domain.

## 2.6 R-WhONet (Recalibrated Wheel Odometry neural Network) Model

This describes the WhONet model which has its conditional probability distribution $P(\boldsymbol{Y}_T|\boldsymbol{X}_T)$ in the target domain (new vehicle), adapted from the source domain $P(\boldsymbol{Y}_S|\boldsymbol{X}_S)$. The conditional probability distribution is formulated from the source vehicle and then adapted to $P(\boldsymbol{Y}_T|\boldsymbol{X}_T)$ in the target domain for deployment on the new target vehicle.
As such, R-WhONet is characterised by the following:
1. $\mathcal{X}_S \neq \mathcal{X}_T$, the feature spaces of the target and source domain are not the same.
2. $P(\boldsymbol{X}_S) \neq P(\boldsymbol{X}_T)$, the marginal probability distributions of the target and source domain are different.
3. $\mathcal{Y}_S \neq \mathcal{Y}_T$, the labels between the source and target tasks vary.
4. $P(\boldsymbol{Y}_S|\boldsymbol{X}_S) \neq P(\boldsymbol{Y}_T|\boldsymbol{X}_T)$, the conditional probability distributions of the target and source tasks differ.

# 3. Experiments and Dataset

## 3.1 Dataset

The WhONet models in this study were trained on both the Inertial and Odometry Vehicle Navigation Benchmark Dataset (IO-VNBD) [45] and Coventry University Public Road dataset for Automated Cars (CUPAC) [46].

### 3.1.1 IO-VNBD

This dataset captured various driving scenarios which across 5700km and 98 hours of driving. These include roundabouts, hard brakes, traffic, sharp turns, etc. on different road types across public roads in the United Kingdom. The data was collected from the Electronic Control Unit (ECU) of a Ford Fiesta Titanium (as shown in Figure 4) at a sampling rate of 10 Hz. The dataset provides information on the vehicle's wheel speeds (in rad/sec), GPS coordinates (in degrees), amongst other sensors describing the motion state of the vehicle. Additional information about the IOVNB dataset can be found in [45].

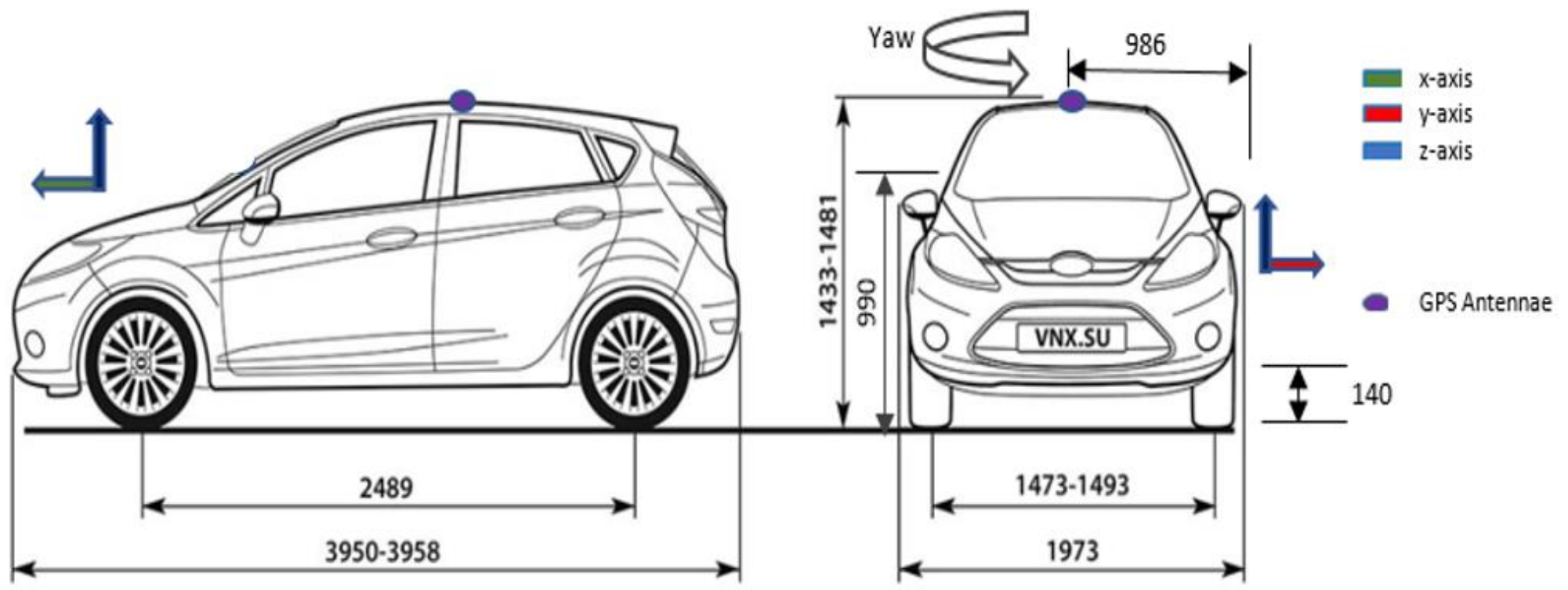

***Figure 4.*** *Data collection vehicle, showing sensor locations* [45].

### 3.1.2 CUPAC

The CUPAC dataset was collected under different road, traffic and weather conditions on public streets in Coventry, United Kingdom. A Ford Fiesta was equipped with a LIDAR unit, a GPS receiver, smartphone sensors, vehicle CAN bus data logger and monocular, infrared and smartphone cameras which recorded data at 10 Hz. These sensors predominately gather data used for vision-based navigation, whilst in this study the wheel speed and GPS positional coordinates of the CAN bus are used. Additional information about the CUPAC dataset can be found in [46].

## 3.2 Performance Metrics

The metrics described as-follows were used to evaluate the performance of the WhONet models and the Wheel encoder Physics Wodel (WPM):

**Cumulative Root Square Error (CRSE):** The CRSE refers to the cumulative root mean squared of the prediction error for every one second of the total duration of the GNSS outage. To provide a better understanding of the performance of the positioning techniques analysed in this study, this value disregards the negative sign of the estimated errors. Equation (16) mathematically describes the CRSE.

$$\text{CRSE} = \sum_{t=1}^{N_t} \sqrt{{e_{pred}}^2} \tag{16}$$

Where $N_t$ is GNSS outage length, $e_{pred}$ refers to the prediction error, and $t$ represents the sampling period which we define as 1 second in this research.

**Cumulative True Error (CTE):** The CTE sums the prediction error from every one second time interval of the total GNSS outage duration. The positive and negative signs of the error estimations are taken into account as no square root value is compiled in this calculation. Over estimations and under estimations of the position errors during the GNSS outage can be better understood with this metric. The CTE is less realistic in comparing the performances of positioning techniques compared to the CRSE, as the negative error predictions could provide a false performance measurement of the positioning technique investigated.

$$\text{CTE} = \sum_{t=1}^{N_t} e_{pred} \tag{17}$$

**Mean (μ):** The mean describes the statistical average of the CRSE and CTE in each GNSS outage scenario, across all the test sequences analysed. The mean of the CRSE and CTE is expressed mathematically in $Equation$ (18),

$$\mu_{CRSE} = \frac{1}{N_s}\sum_{i=1}^{N_s} CRSE, \mu_{CTE} = \frac{1}{N_s}\sum_{i=1}^{N_s} CTE \tag{18}$$

where $N_s$ stands for the total number of test sequences in each scenario.

**Standard deviation (*σ*)**: Standard deviation measures the variation and spread of CRSE and CTE results of all test sequences in each scenario investigated.

$$\sigma_{CRSE} = \sqrt{\frac{\sum(CRSE_i - \mu_{CRSE})^2}{N_s}}, \sigma_{CTE} = \sqrt{\frac{\sum(CTE_i - \mu_{CTE})^2}{N_s}} \tag{19}$$

**Minimum (Min)**: This value describes the minimum CTE and CRSE of all the test sequences investigated in each GNSS scenario.

**Maximum (Max)**: This value describes the maximum CTE and CRSE of all test sequences analysed in each GNSS scenario.

## 3.3 Model Training

This section details how the G-WhONet, S-WhONet and R-WhONet models are trained using the Keras-Tensorflow platform.

### 3.3.1 G-WhONet's Training

G-WhONet is trained on the IO-VNB data subsets presented in Table A1 (see the appendix) just as the WhONet model in [11], thus establishing the conditional distribution $P(y_s|X_s)$. The IO-VNB datasets used to train the G-WhONet Model are characterised by about 1590 mins of drive time over a total distance of 1,165 km. The G-WhONet model was optimised using the adamax optimiser with an initial learning rate of 0.0007 and trained using the mean absolute error loss function and a dropout rate of 0.05 of the weighted connections in the hidden layers. Furthermore, the input features to the G-WhONet were normalised between 0 and 1 to reduce learning bias. Table 1 highlights the parameters characterising the training of the G-WhONet model.

***Table 11.*** *Training parameters of the G-WhONet, S-WhONet and R-WhONet models.*

| Parameters | G-WhONet/ S-WhONet/ R-WhONet |
|---|---|
| Learning rate | 0.0007 |
| Dropout rate | 0.05 |
| Time step | 1 |
| Hidden layers | 1 |
| Hidden neurons | 72 per layer |
| Batch size | 128 |

### 3.3.2 S-WhONet's Training

S-WhONet was trained on the CUPAC data subsets presented in Table A2 to determine the conditional probability distribution $P(y_T|X_T)$. The CUPAC datasets used to train the S-WhONet model are characterised by about 104 mins of driving over 54.9 km. The S-WhONet model was optimised using the adamax optimiser with an initial learning rate of 0.0007 and trained using the mean absolute error loss function and a dropout rate of 0.05 of the weighted connections in the hidden layers. Just as with the G-WhONet model, the input features to the S-WhONet were normalised between 0 and 1 to reduce learning bias. Table 1 highlights the parameters characterising the training of the S-WhONet model.

### 3.3.3 R-WhONet's Training

The G-WhONet model of the source domain with conditional probability distribution $P(y_S|X_S)$ is adapted to create the R-WhONet model which learns the target probability distribution $P(y_T|X_T)$ of the new vehicle on which it is to be deployed. The R-WhONet model is trained on the first $m$ seconds of the CUPAC datasets described in Table A3. The input features to the R-WhONet model were normalised between 0 and 1 to reduce learning bias. Table 1 highlights the parameters characterising the training (adaptation) of the R-WhONet model. Figure 5 describes the learning relationship between the R-WhOnet and WhONet as well as the RWhONet's learning scheme.

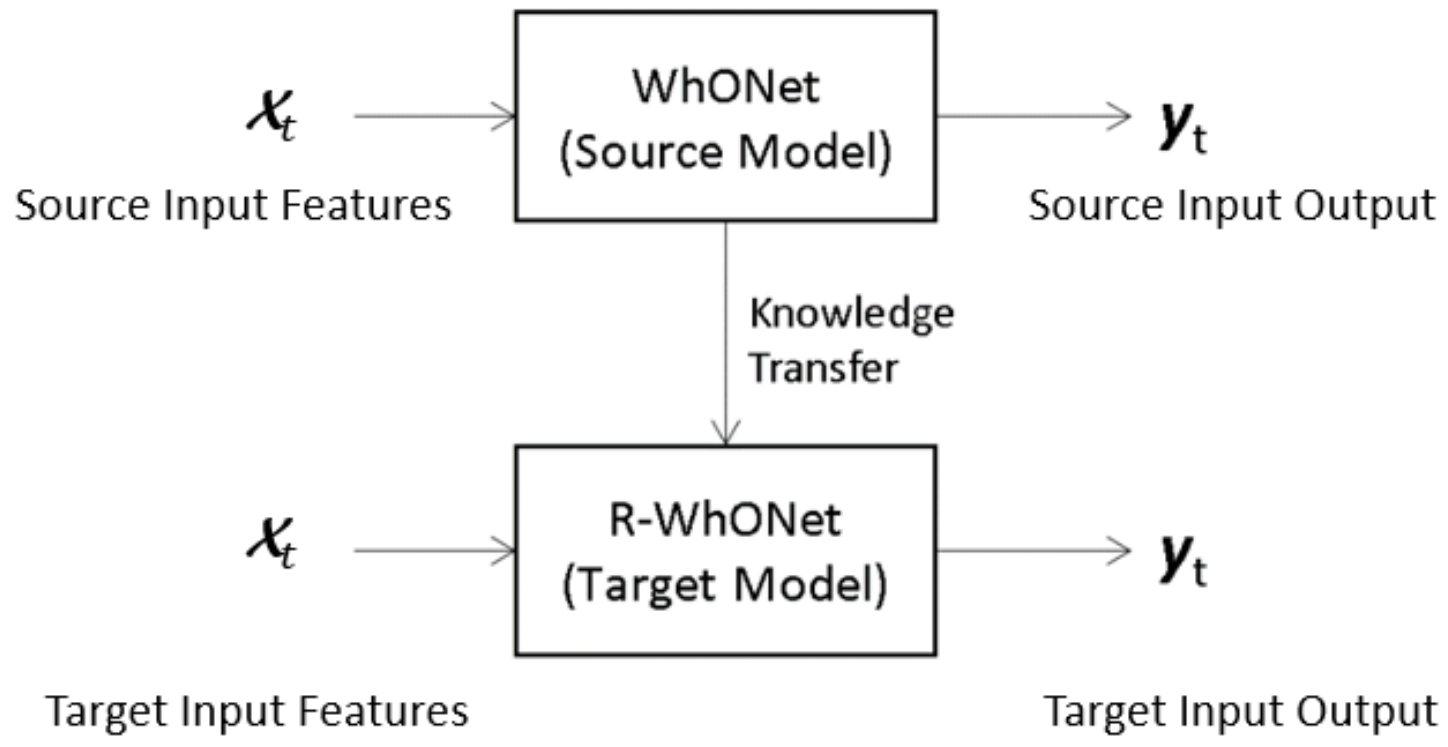

***Figure 5.*** *Learning scheme of the R-WhONet model*

## 3.4 R-WhONet's Training Time Selection Experiment

The R-WhONet model is trained on the first $m$ seconds of the CUPAC datasets presented on Table A3. In order to determine the value of $m$ that consistently gives the least position uncertainty estimation $y_T$ in the target domain, we experimented over a range of values for m (10 s, 20 s, 30 s,…., and 100 s). The results from this experimentation are presented in Table 2. From observation, it can be seen that when $m$ is defined as 50 seconds, R-WhONet provides the least position error estimation when averaged. This shows that the R-WhONet needs 50 seconds of data before it is able to most accurately adapt to its target domain. At an $m$ value of 40 seconds, slightly worse results were obtained compared to an $m$ value of 50 seconds, however at all other values of $m$, very poor results were obtained. This could be possibly attributed to the R-WhONet underfitting or overfitting the data in the target domain.

***Table 2.*** *2Results from the experimentation of various values of m to determine the adaptation time for the R-WhONet model.*

| CUPAC | Positional Cumulative Root Square Error (CRSE) (m) | | | | | | | | | | |
|---|---|---|---|---|---|---|---|---|---|---|---|
| Dataset | m | 10 s | 20 s | 30 s | 40 s | 50 s | 60 s | 70 s | 80 s | 90 s | 100 s |
| Charlie 3 | Max | 14.8 | 243.8 | 68.2 | 11.2 | 10.6 | 124.5 | 75.5 | 69.7 | 148.5 | 33.7 |
| | Min | 5.6 | 7.2 | 6.7 | 4.5 | 3.5 | 6.3 | 6.3 | 6.5 | 6.8 | 6.5 |
| | (μ) | 10.1 | 66.0 | 21.6 | 9.0 | 7.8 | 30.0 | 23.0 | 22.9 | 48.5 | 18.8 |
| | (σ) | 2.0 | 81.1 | 20.2 | 1.0 | 1.4 | 39.0 | 21.9 | 20.0 | 51.4 | 9.7 |
| Charlie 4 | **Max** | 21.5 | 29.8 | 21.8 | 21.5 | 21.1 | 21.6 | 22.8 | 23.4 | 23.5 | 23.9 |
| | **Min** | 5.5 | 5.1 | 5.5 | 5.5 | 5.6 | 5.7 | 6.4 | 6.8 | 8.2 | 7.3 |
| | **(μ)** | 11.5 | 15.0 | 11.8 | 11.6 | 11.5 | 12.1 | 12.5 | 13.4 | 15.9 | 13.4 |
| | **(σ)** | 3.8 | 5.1 | 5.9 | 4.1 | 4.0 | 5.2 | 5.1 | 5.1 | 5.0 | 6.1 |
| Delta 2 | **Max** | 7.0 | 7.1 | 8.0 | 7.7 | 7.3 | 6.8 | 6.8 | 7.1 | 7.2 | 7.4 |
| | **Min** | 6.8 | 6.4 | 7.4 | 7.0 | 6.6 | 6.4 | 6.7 | 6.8 | 6.5 | 6.7 |
| | **(μ)** | 7.0 | 6.9 | 7.7 | 7.5 | 7.1 | 6.6 | 6.7 | 7.0 | 6.9 | 7.1 |
| | **(σ)** | 0.0 | 0.0 | 0.0 | 0.0 | 0.0 | 0.0 | 0.0 | 0.0 | 0.1 | 0.0 |
| Delta 3 | **Max** | 11.9 | 8.7 | 10.9 | 8.7 | 10.2 | 11.8 | 12.1 | 12.5 | 7.8 | 8.1 |
| | **Min** | 7.3 | 6.9 | 7.4 | 6.7 | 6.4 | 6.6 | 6.0 | 5.8 | 5.9 | 6.2 |
| | **(μ)** | 9.4 | 7.9 | 8.9 | 7.5 | 8.1 | 8.3 | 8.0 | 7.7 | 6.7 | 7.0 |
| | **(σ)** | 0.9 | 0.4 | 0.7 | 0.6 | 0.8 | 1.2 | 0.8 | 1.1 | 0.6 | 0.5 |

## 3.5 Model Evaluation

The R-WhONet, G-WhONet and S-WhONet and the physics model (WPM) are evaluated on the CUPAC datasets presented in Table A3 The CUPAC dataset used for evaluation has vehicular dynamics and state properties such as tyre pressure, worn-out tyre state, driving behaviour, etc. different from the IO-VNBD dataset which was used for training the G-WhONet model. The models are evaluated on the longer-term GNSS outage scenarios of 30s, 60s, 120s and 180s just as in [11].

Each test set used in the longer-term GNSS outage scenarios are broken down into test sequences of 30s, 60s, 120s or 180s depending on the outage scenario being evaluated. The maximum, minimum, average and standard deviation, of the CRSE's and CTE's of all the test sequences evaluated within each dataset are recorded in each scenario and used to evaluate each model's performance. GPS outages are assumed on the test scenarios for the purpose of the investigation with a prediction frequency of 1s.

## 4. Results and Discussion

In this section, the performance of the R-WhONet in comparison to the G-WhONet model, S-WhONet model and WPM are evaluated on the 30 s, 60 s, 120 s and 180 s scenarios using the CUPAC datasets presented in Table A3.

## 4.1 30s GNSS Outage

In the 30 second GNSS outage scenario, R-WhONet provides a CRSE reduction of up to 28% on the G-WhONet's estimation. As presented in Table 3 and 4, the R-WhONet achieves the best average CTE and CRSE of 0.31 m and 1.22 m compared to 1.11 m and 1.46 m of the G-WhONet over a max distance per test sequence of 618 m of all 121 sequences evaluated on the CUPAC dataset (see Table A3). Moreover, on the maximum metric, the R-WhONet achieves the best CTE and CRSE of 0.92 m and 1.60 m respectively while G-WhONet achieves the best CTE and CRSE of 1.96 m and 2.02 m respectively as well.

The robustness of the R-WhONet is further emphasized by the low standard deviation of 0.20 obtained compared to 0.32 of G-WhONet. The result so obtained shows that R-WhONet is able to adapt the G-WhONet model to the new vehicle with performances relatively closer to a model (S-WhONet) trained specifically for the new vehicle as further illustrated in Figures 5 and 6.

***Table 3****. CRSE performance comparison of the R-WhONet, G-WhONet, S-WhONet and WPM on the 30 seconds GNSS outage scenario.*

| **CUPAC Dataset** | **Performance Metrics** | **Positional Estimation Error (m)** | | | | **Total Distance Travelled / (m)** | **Number of Test Sequences evaluated** |
|---|---|---|---|---|---|---|---|
| | | **Cumulative Root Square Error (CRSE)** | | | | | |
| | | **WPM** | **R-WhONet** | **G-WhONet** | **S-WhONet** | | |
| Charlie 3 | **Max** | 13.71 | 4.14 | 5.68 | 5.36 | 618 | 45 |
| | **Min** | 2.25 | 0.50 | 0.56 | 0.37 | 18 | |
| | **(μ)** | 7.43 | 1.34 | 1.61 | 1.14 | 313 | |
| | **(σ)** | 2.81 | 0.71 | 0.90 | 0.73 | 154 | |
| Charlie 4 | **Max** | 16.02 | 6.25 | 9.70 | 8.53 | 407 | 22 |
| | **Min** | 2.95 | 0.62 | 0.80 | 0.41 | 12 | |
| | **(μ)** | 7.42 | 1.89 | 2.05 | 1.62 | 217 | |
| | **(σ)** | 2.79 | 1.35 | 2.00 | 1.69 | 95 | |
| Delta 2 | **Max** | 23.26 | 1.60 | 2.02 | 1.42 | 261 | 16 |
| | **Min** | 3.19 | 0.83 | 0.69 | 0.51 | 55 | |
| | **(μ)** | 8.85 | 1.22 | 1.46 | 0.94 | 193 | |
| | **(σ)** | 4.72 | 0.20 | 0.32 | 0.24 | 65 | |
| Delta 3 | **Max** | 12.45 | 3.74 | 5.22 | 5.07 | 346 | 38 |
| | **Min** | 4.05 | 0.84 | 0.90 | 0.43 | 8 | |
| | **(μ)** | 7.22 | 1.53 | 1.77 | 1.18 | 131 | |
| | **(σ)** | 2.47 | 0.57 | 0.70 | 0.72 | 97 | |

***Table 4.*** *CTE performance comparison of the R-WhONet, G-WhONet, S-WhONet and WPM on the 30 seconds GNSS outage scenario.*

| **CUPAC Dataset** | **Performance Metrics** | **Positional Estimation Error (m)** | | | | **Total Distance Travelled (m)** | **Number of Test Sequences evaluated** |
|---|---|---|---|---|---|---|---|
| | | **Cumulative True Error (CTE)** | | | | | |
| | | **WPM** | **R-WhONet** | **G-WhONet** | **S-WhONet** | | |
| Charlie 3 | **Max** | 10.26 | 2.90 | 3.90 | 2.05 | 618 | 45 |
| | **Min** | 0.18 | 0.00 | 0.03 | 0.00 | 18 | |
| | **(μ)** | 2.96 | 0.60 | 1.13 | 0.46 | 313 | |
| | **(σ)** | 2.15 | 0.61 | 0.93 | 0.47 | 154 | |
| Charlie 4 | **Max** | 5.95 | 3.30 | 6.42 | 3.87 | 407 | 22 |
| | **Min** | 0.61 | 0.00 | 0.34 | 0.00 | 12 | |
| | **(μ)** | 2.82 | 0.84 | 1.46 | 0.81 | 217 | |
| | **(σ)** | 1.70 | 0.83 | 1.37 | 0.86 | 95 | |
| Delta 2 | **Max** | 4.24 | 0.92 | 1.96 | 0.78 | 261 | 16 |
| | **Min** | 0.83 | 0.00 | 0.16 | 0.01 | 55 | |
| | **(μ)** | 2.61 | 0.31 | 1.11 | 0.30 | 193 | |
| | **(σ)** | 1.21 | 0.27 | 0.42 | 0.20 | 65 | |
| Delta 3 | **Max** | 7.58 | 1.88 | 2.47 | 1.36 | 346 | 38 |
| | **Min** | 0.19 | 0.00 | 0.09 | 0.00 | 8 | |
| | **(μ)** | 2.08 | 0.55 | 1.33 | 0.37 | 131 | |
| | **(σ)** | 1.85 | 0.45 | 0.52 | 0.33 | 97 | |

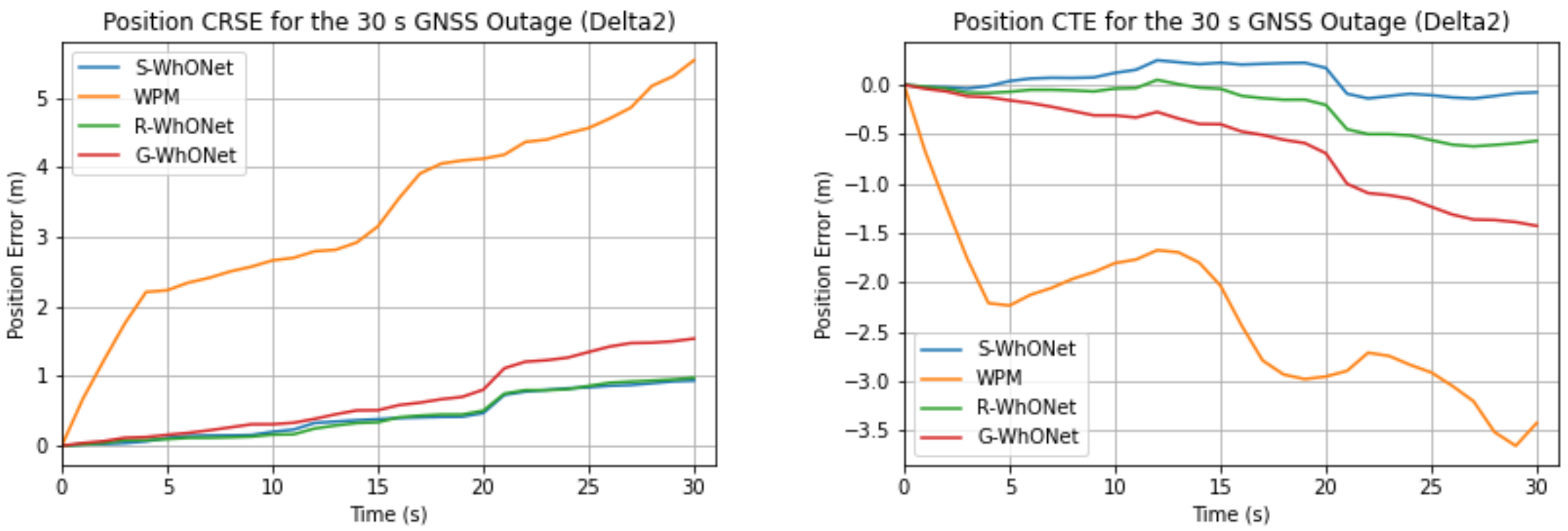

***Figure 51.*** *Showing a sample* ***Cumulative Root Square Error (CRSE)*** *and* ***Cumulative True Error (CTE)*** *evolution of the models on the Delta 2 CUPAC dataset during the 30 s GNSS outage scenario after recalibration.*

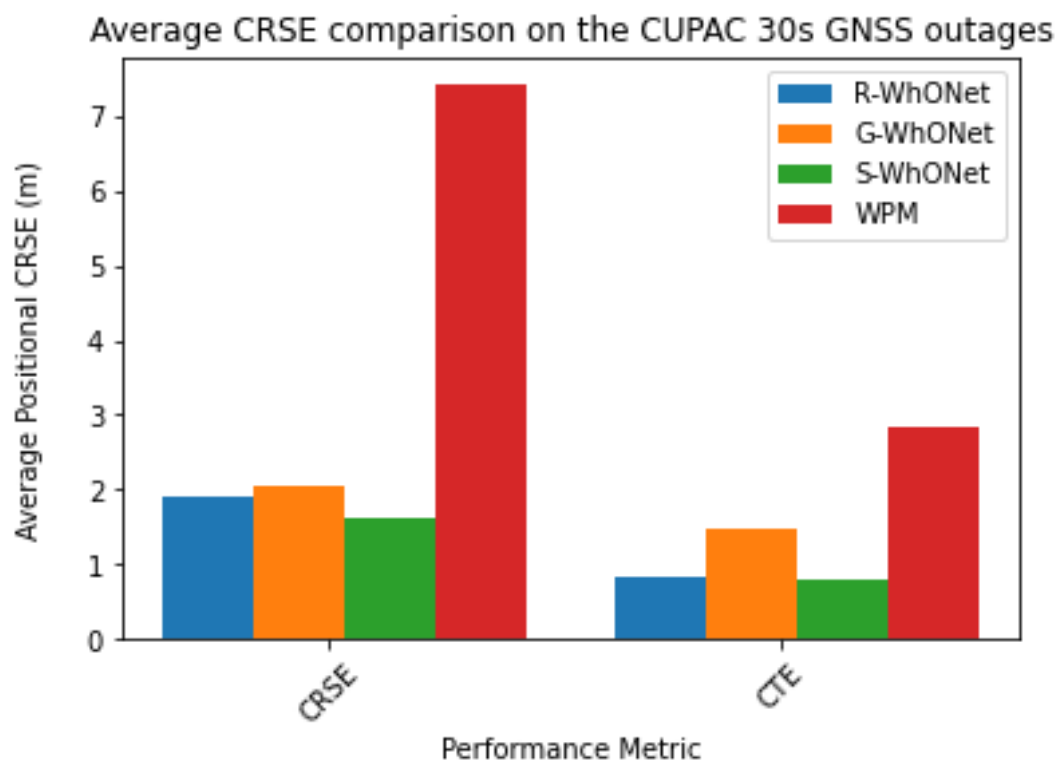

***Figure 6.2** **Cumulative Root Square Error (CRSE)** and **Cumulative True Error (CTE)** comparison of the models on the Charlie 4 CUPAC dataset during the 30 s GNSS outage scenario after recalibration.*

## 4.2 60s GNSS Outage

Tables 5 and 6 report the results from the 60 seconds GNSS-outage CUPAC experiment. On the CTE and CRSE metric, the R-WhONet provides the best average of 0.50 m and 2.45 m respectively, compared to 2.15 m and 2.89 m respectively for the G-WhONet over a max test sequence distance covered of 1234 m of all 60 sequences evaluated on the CUPAC dataset (see Table A3). Also, on the maximum metric, the R-WhONet obtains the best CTE and CRSE of 0.83 m and 2.72 m in contrast to the G-WhONet best CTE and CRSE of 3.02 m and 3.65 m.

The consistency of the R-WhONet is further highlighted by the low standard deviation of 0.18 obtained compared to 0.42 of G-WhONet as reported in Table 5. The evolution of the R-WhONet, G-WhONet, S-WhONet and WPM error estimations over time on both the CTE and CRSE metrics as well as the error comparisons are illustrated in Figures 7 and 8.

***Table 5.** **Cumulative Root Square Error (CRSE)** performance comparison of the R-WhONet, G-WhONet, S-WhONet and WPM on the 60 seconds GNSS outage scenario.*

| CUPAC Dataset | Performance Metrics | Positional Estimation Error (m) Cumulative Root Square Error (CRSE) WPM | R-WhONet | G-WhONet | S-WhONet | Total Distance Travelled (m) | Number of Test Sequences evaluated |
|---|---|---|---|---|---|---|---|
| Charlie 3 | **Max** | 21.48 | 5.56 | 6.62 | 4.94 | 1234 | 22 |
| | **Min** | 6.08 | 1.13 | 1.43 | 1.03 | 281 | |
| | **(μ)** | 14.86 | 2.75 | 3.25 | 2.54 | 627 | |
| | **(σ)** | 4.14 | 0.94 | 1.32 | 0.90 | 284 | |
| Charlie 4 | **Max** | 19.40 | 11.26 | 12.46 | 11.41 | 643 | 11 |
| | **Min** | 7.91 | 1.63 | 2.21 | 1.50 | 250 | |
| | **(μ)** | 15.06 | 3.76 | 4.18 | 3.70 | 445 | |
| | **(σ)** | 3.51 | 1.87 | 3.13 | 2.34 | 149 | |
| Delta 2 | **Max** | 32.17 | 2.72 | 3.65 | 2.60 | 490 | 8 |
| | **Min** | 8.41 | 2.02 | 2.15 | 1.66 | 149 | |
| | **(μ)** | 17.69 | 2.45 | 2.89 | 2.22 | 385 | |
| | **(σ)** | 7.01 | 0.18 | 0.47 | 0.27 | 104 | |
| Delta 3 | **Max** | 18.19 | 5.00 | 4.31 | 4.23 | 661 | 19 |
| | **Min** | 10.17 | 2.00 | 2.76 | 1.56 | 24 | |
| | **(μ)** | 14.45 | 2.76 | 3.36 | 2.46 | 253 | |
| | **(σ)** | 2.74 | 0.63 | 0.42 | 0.45 | 168 | |

***Table 6.** CTE performance comparison of the R-WhONet, G-WhONet, S-WhONet and WPM on the 60 seconds GNSS outage scenario.*

| CUPAC Dataset | Performance Metrics | Positional Estimation Error (m) Cumulative True Error (CTE) WPM | R-WhONet | G-WhONet | S-WhONet | Total Distance Travelled (m) | Number of Test Sequences evaluated |
|---|---|---|---|---|---|---|---|
| Charlie 3 | **Max** | 9.17 | 2.90 | 6.59 | 2.61 | 1234 | 22 |
| | **Min** | 0.10 | 0.00 | 0.11 | 0.00 | 281 | |
| | **(μ)** | 3.85 | 1.03 | 2.15 | 0.81 | 627 | |
| | **(σ)** | 2.56 | 0.81 | 1.53 | 0.73 | 284 | |
| Charlie 4 | **Max** | 8.44 | 3.45 | 6.87 | 3.96 | 643 | 11 |
| | **Min** | 1.35 | 0.00 | 0.00 | 0.00 | 250 | |
| | **(μ)** | 3.94 | 1.37 | 2.71 | 1.41 | 445 | |
| | **(σ)** | 2.04 | 0.68 | 2.06 | 0.84 | 149 | |
| Delta 2 | **Max** | 6.14 | 0.83 | 3.02 | 0.75 | 490 | 8 |
| | **Min** | 0.67 | 0.00 | 1.30 | 0.01 | 149 | |
| | **(μ)** | 3.32 | 0.50 | 2.21 | 0.50 | 385 | |
| | **(σ)** | 1.95 | 0.25 | 0.63 | 0.14 | 104 | |
| Delta 3 | **Max** | 9.89 | 2.92 | 4.22 | 1.49 | 661 | 19 |
| | **Min** | 0.01 | 0.02 | 0.21 | 0.00 | 24 | |
| | **(μ)** | 2.45 | 0.80 | 2.59 | 0.52 | 253 | |
| | **(σ)** | 2.36 | 0.69 | 0.82 | 0.40 | 168 | |

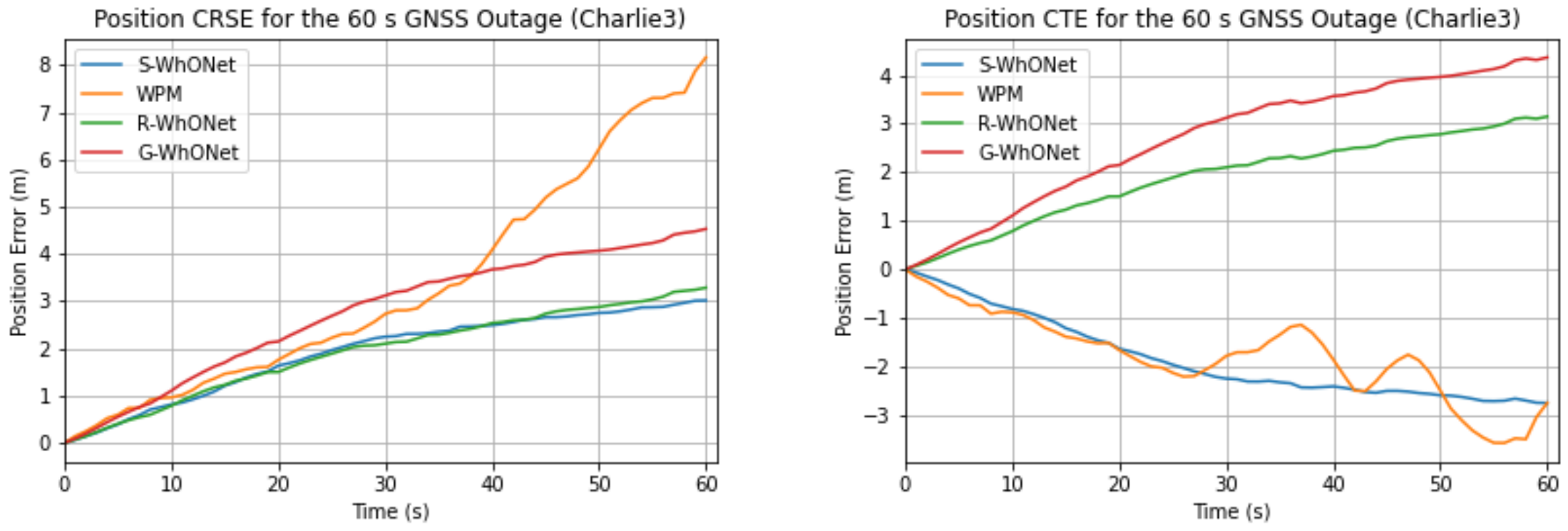

*Figure 73. Showing a sample* ***Cumulative Root Square Error (CRSE)*** *and* ***Cumulative True Error (CTE)*** *evolution of the models on the Delta 2 CUPAC dataset during the 60 s GNSS outage scenario after recalibration.*

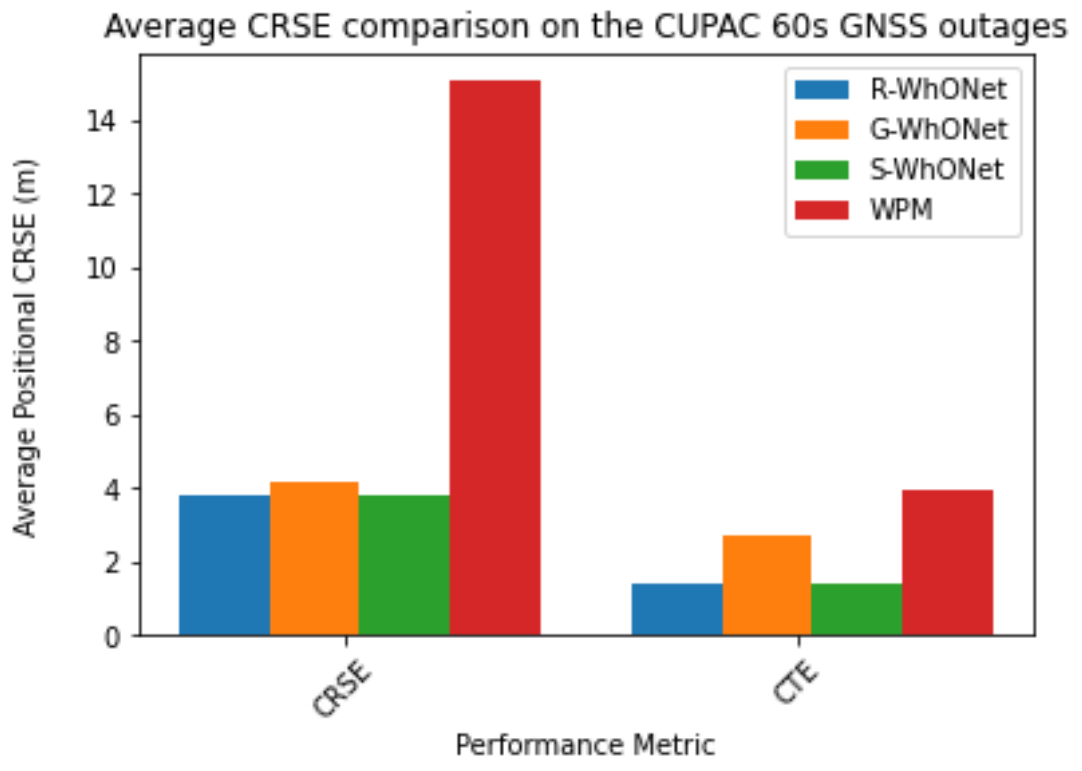

***Figure 84.*** ***Cumulative Root Square Error (CRSE)*** *and* ***Cumulative True Error (CTE)*** *comparison of the models on the Charlie 4 CUPAC dataset during the 60 s GNSS outage scenario after recalibration.*

## 4.3 120s GNSS Outage

The R-WhONet outperforms the G-WhONet on the CRSE metric by up to 31% on the 120 seconds GNSS outage scenario. When evaluated on the CUPAC dataset, as shown in Table 7 and 8, the R-WhONet achieves the best average CTE and CRSE of 0.68 m and 4.67 m compared to 4.09 m and 5.54 m for the G-WhONet over a maximum distance per test sequence of 2262 m. In addition, on the maximum metric, the R-WhONet has the best CTE and CRSE of 0.99 m and 4.75 m, while G-WhONet has CTE and CRSE of 5.62 m and 6.26 m respectively.

The consistency of the R-WhONet is further emphasised by the low standard deviation of 0.02 obtained in comparison to 0.52 for G-WhONet as reported in Table 7. The result so obtained shows that R-WhONet is able to adapt the G-WhONet model to the new vehicle with performances relatively close to a model (S-WhONet) trained specifically for the new vehicle as further illustrated in Figures 9 and 10.

***Table 7.*** *CRSE performance comparison of the R-WhONet, G-WhONet, S-WhONet and WPM on the 120 seconds GNSS outage scenario.*

| **CUPAC Dataset** | **Performance Metrics** | **Positional Estimation Error (m) Cumulative Root Square Error (CRSE)** | | | | **Total Distance Travelled (m)** | **Number of Test Sequences evaluated** |
|---|---|---|---|---|---|---|---|
| | | **WPM** | **R-WhONet** | **G-WhONet** | **S-WhONet** | | |
| Charlie 3 | **Max** | 38.45 | 10.32 | 9.44 | 9.40 | 2262 | 11 |
| | **Min** | 15.64 | 3.45 | 4.07 | 2.27 | 637 | |
| | **(μ)** | 29.71 | 5.69 | 6.49 | 4.57 | 1254 | |
| | **(σ)** | 7.53 | 1.46 | 1.71 | 1.61 | 539 | |
| Charlie 4 | **Max** | 38.54 | 18.25 | 19.82 | 12.10 | 1140 | 5 |
| | **Min** | 25.10 | 3.53 | 4.44 | 2.60 | 558 | |
| | **(μ)** | 30.12 | 7.86 | 8.36 | 6.65 | 890 | |
| | **(σ)** | 4.77 | 3.54 | 5.79 | 3.71 | 204 | |
| Delta 2 | **Max** | 47.53 | 4.75 | 6.26 | 4.68 | 887 | 4 |
| | **Min** | 20.76 | 4.45 | 5.02 | 2.70 | 540 | |
| | **(μ)** | 35.90 | 4.67 | 5.54 | 3.75 | 735 | |
| | **(σ)** | 11.21 | 0.02 | 0.52 | 0.45 | 145 | |
| Delta 3 | **Max** | 33.37 | 5.93 | 7.64 | 5.74 | 1165 | 9 |
| | **Min** | 24.09 | 3.90 | 5.65 | 3.15 | 204 | |
| | **(μ)** | 29.34 | 5.17 | 6.73 | 4.30 | 505 | |
| | **(σ)** | 3.88 | 0.39 | 0.62 | 0.66 | 310 | |

***Table 8****. performance comparison of the R-WhONet, G-WhONet, S-WhONet and WPM on the 120 seconds GNSS outage scenario.*

| **CUPAC Dataset** | **Performance Metrics** | **Positional Estimation Error (m) Cumulative True Error (CTE)** | | | | **Total Distance Travelled (m)** | **Number of Test Sequences evaluated** |
|---|---|---|---|---|---|---|---|
| | | **WPM** | **R-WhONet** | **G-WhONet** | **S-WhONet** | | |
| Charlie 3 | **Max** | 14.78 | 2.86 | 8.50 | 2.34 | 2262 | 11 |
| | **Min** | 0.14 | 0.00 | 1.09 | 0.00 | 637 | |
| | **(μ)** | 6.39 | 1.40 | 4.09 | 1.06 | 1254 | |
| | **(σ)** | 4.43 | 0.94 | 2.23 | 0.72 | 539 | |
| Charlie 4 | **Max** | 12.93 | 5.40 | 12.89 | 5.48 | 1140 | 5 |
| | **Min** | 0.29 | 0.00 | 1.50 | 0.01 | 558 | |
| | **(μ)** | 5.32 | 2.46 | 5.41 | 2.64 | 890 | |
| | **(σ)** | 4.53 | 0.55 | 3.94 | 0.82 | 204 | |
| Delta 2 | **Max** | 7.63 | 0.99 | 5.62 | 1.08 | 887 | 4 |
| | **Min** | 1.51 | 0.07 | 3.08 | 0.01 | 540 | |
| | **(μ)** | 4.37 | 0.68 | 4.15 | 0.79 | 735 | |
| | **(σ)** | 2.52 | 0.35 | 1.08 | 0.08 | 145 | |
| Delta 3 | **Max** | 9.82 | 2.71 | 6.84 | 1.49 | 1165 | 9 |
| | **Min** | 0.65 | 0.00 | 3.01 | 0.02 | 204 | |
| | **(μ)** | 3.84 | 1.34 | 5.19 | 0.83 | 505 | |
| | **(σ)** | 3.15 | 0.83 | 1.20 | 0.40 | 310 | |

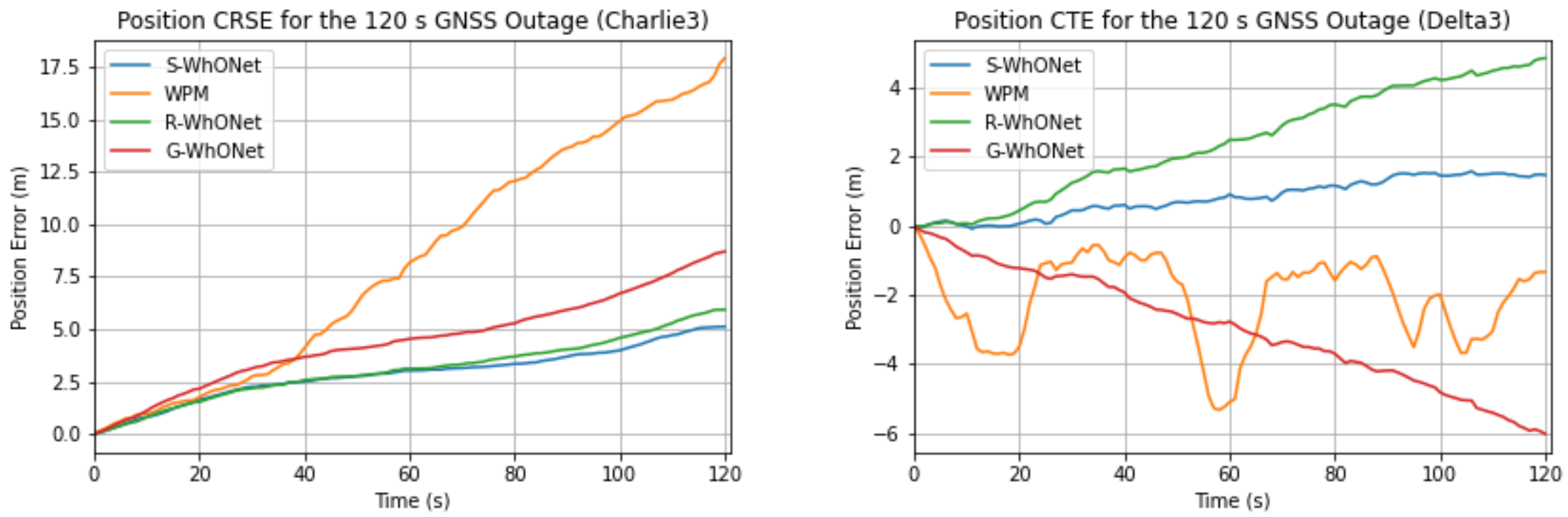

***Figure 9.5** Showing a sample **Cumulative Root Square Error** (CRSE) and **Cumulative True Error (CTE)** evolution of the models on the Delta 2 CUPAC dataset during the 120 s GNSS outage scenario after recalibration.*

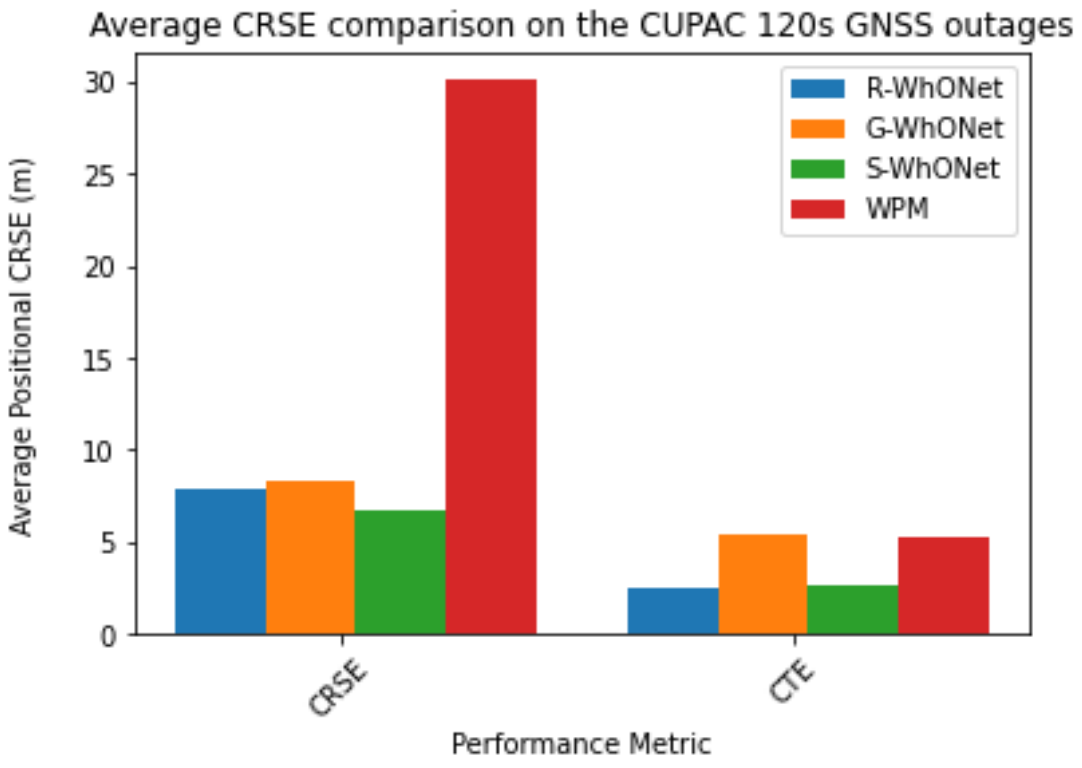

***Figure 106.** **Cumulative Root Square Error (CRSE)** and **Cumulative True Error (CTE)** comparison of the models on the Charlie 4 CUPAC dataset during the 120 s GNSS outage scenario after recalibration.*

## 4.4 180s GNSS Outage

In the 180 second GNSS outage scenario, R-WhONet provides a CRSE reduction of up to 32% on the G-WhONet's estimation. As presented on Tables 9 and 10, the R-WhONet achieves the best average CTE and CRSE of 0.26 m and 6.84 m compared to 5.50 m and 8.31 m of the G-WhONet over a max distance per test sequence of 3496 m of all 18 sequences evaluated on the CUPAC dataset (s). Furthermore, on the maximum metric, the R-WhONet achieves the best CTE and CRSE of 0.49 m and 7.30 m compared to the G-WhONet's best CTE and CRSE of 6.46 m and 8.43 m.

The consistency of the R-WhONet is better accentuated by the low standard deviation of 0 obtained, compared to 0.12 of G-WhONet, as reported in Tables 9 and 10. The result so obtained shows that R-WhONet is able to adapt the G-WhONet model to the new vehicle domain, with performances relatively closer to a model (S-WhONet) trained specifically for the new vehicle as further illustrated in Figures 11 and 12.

***Table 9.** CRSE performance comparison of the R-WhONet, G-WhONet, S-WhONet and WPM on the 180 seconds GNSS outage scenario.*

| CUPAC Dataset | Performance Metrics | Positional Estimation Error (m) Cumulative Root Square Error (CRSE) | | | | Total Distance Travelled (m) | Number of Test Sequences evaluated |
|---|---|---|---|---|---|---|---|
| | | WPM | R-WhONet | G-WhONet | S-WhONet | | |
| Charlie 3 | **Max** | 55.25 | 10.34 | 15.31 | 9.94 | 3496 | 7 |
| | **Min** | 23.99 | 4.62 | 5.49 | 3.74 | 1012 | |
| | **(μ)** | 44.42 | 7.87 | 9.86 | 7.69 | 1916 | |
| | **(σ)** | 9.19 | 1.38 | 2.90 | 1.31 | 772 | |
| Charlie 4 | **Max** | 54.72 | 21.23 | 22.44 | 21.14 | 1716 | 3 |
| | **Min** | 41.51 | 5.50 | 7.36 | 5.22 | 1220 | |
| | **(μ)** | 46.94 | 11.45 | 12.52 | 11.28 | 1400 | |
| | **(σ)** | 5.64 | 4.24 | 7.01 | 5.27 | 224 | |
| Delta 2 | **Max** | 68.59 | 7.30 | 8.43 | 6.58 | 1278 | 2 |
| | **Min** | 39.12 | 6.22 | 8.19 | 6.15 | 928 | |
| | (μ) | 53.86 | 6.84 | 8.31 | 6.39 | 1103 | |
| | (σ) | 14.74 | 0.04 | 0.12 | 0.00 | 175 | |
| Delta 3 | Max | 48.63 | 10.18 | 10.52 | 9.20 | 1624 | 6 |
| | Min | 39.39 | 6.35 | 9.06 | 6.10 | 323 | |
| | (μ) | 44.01 | 7.73 | 10.09 | 7.38 | 758 | |
| | (σ) | 4.62 | 0.94 | 0.51 | 0.50 | 470 | |

***Table 10.*** *CTE performance comparison of the R-WhONet, G-WhONet, S-WhONet and WPM on the 180 seconds GNSS outage scenario.*

| CUPAC Dataset | Performance Metrics | Positional Estimation Error (m) *Cumulative True Error (CTE)* WPM | R-WhONet | G-WhONet | S-WhONet | Total Distance Travelled (m) | Number of Test Sequences evaluated |
|---|---|---|---|---|---|---|---|
| Charlie 3 | Max | 20.66 | 4.84 | 15.09 | 3.35 | 3496 | 7 |
| | Min | 0.48 | 0.05 | 0.98 | 0.03 | 1012 | |
| | (μ) | 10.12 | 1.66 | 5.50 | 1.57 | 1916 | |
| | (σ) | 6.00 | 1.61 | 4.54 | 0.73 | 772 | |
| Charlie 4 | Max | 11.94 | 5.10 | 14.59 | 5.21 | 1716 | 3 |
| | Min | 8.13 | 0.04 | 3.26 | 0.14 | 1220 | |
| | (μ) | 10.14 | 3.60 | 7.78 | 3.68 | 1400 | |
| | (σ) | 1.56 | 0.31 | 4.90 | 0.18 | 224 | |
| Delta 2 | Max | 9.81 | 0.49 | 6.46 | 0.25 | 1278 | 2 |
| | Min | 3.30 | 0.02 | 5.99 | 0.07 | 928 | |
| | (μ) | 6.56 | 0.26 | 6.22 | 0.16 | 1103 | |
| | (σ) | 3.26 | 0.00 | 0.23 | 0.01 | 175 | |
| Delta 3 | Max | 8.73 | 3.17 | 9.54 | 1.10 | 1624 | 6 |
| | Min | 0.46 | 0.03 | 5.43 | 0.00 | 323 | |
| | (μ) | 4.90 | 1.29 | 7.78 | 0.53 | 131 | |
| | (σ) | 3.17 | 0.65 | 1.29 | 0.33 | 97 | |

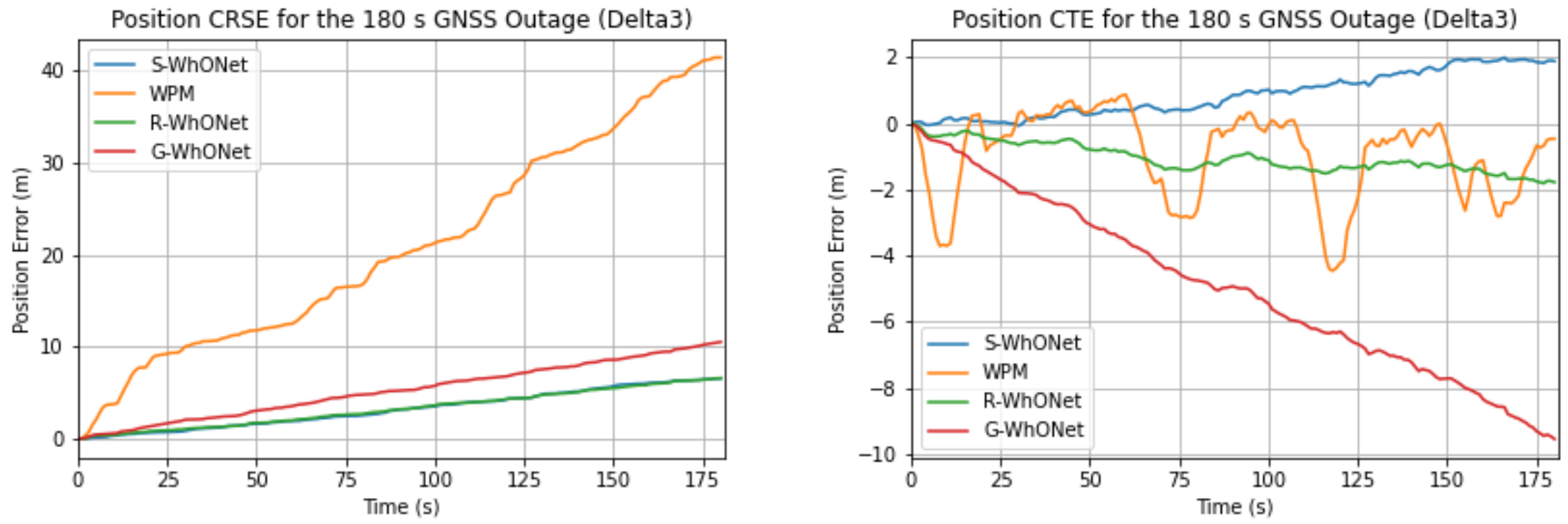

***Figure 117.*** *Showing a sample* ***Cumulative Root Square Error (CRSE)*** *and* ***Cumulative True Error (CTE)*** *evolution of the models on the Delta 2 CUPAC dataset during the 180 s GNSS outage scenario after recalibration.*

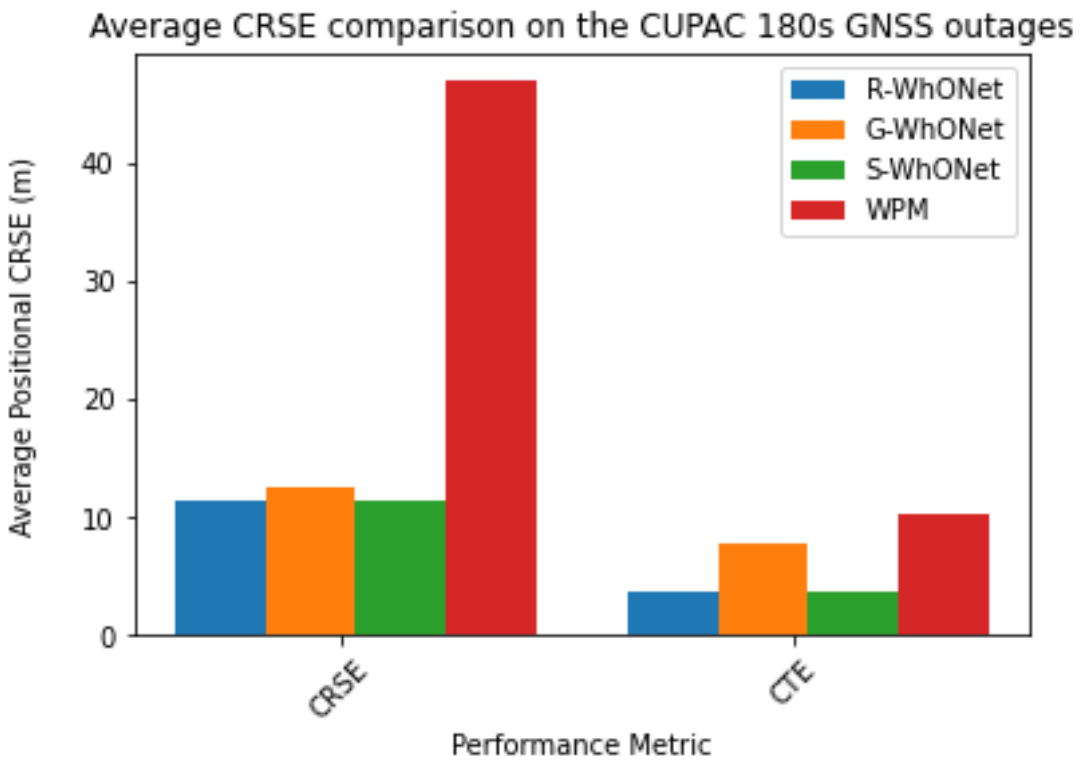

***Figure 128.*** ***Cumulative Root Square Error (CRSE)*** *and* ***Cumulative True Error (CTE)*** *comparison of the models on the Charlie 4 CUPAC dataset during the 180 s GNSS outage scenario after recalibration.*

## 5. Conclusion

In this paper, the R-WhONet (Recalibration Wheel Odometry neural Network) methodology is proposed for the adaptation of the WhONet model to vehicles with different dynamics and states such as tyre pressures, worn-out tyre state, driving behaviour, size, etc. Due to the difference in the feature space of different vehicles as affected by varying vehicle dynamics and state properties, a model trained to learn the correlation between the wheel speed and the position uncertainty on a vehicle A, will not fit a vehicle B with different motion characteristics. As it is expensive and impractical to train a Specific WhONet (S-WhONet) model for every vehicle's dynamics and different worn-out tyre states, as well as tyre pressure, we leveraged transfer learning to recalibrate a Generic WhONet (G-WhONet) model trained on a source Vehicle A, for an on-the-fly adaptation to a new target vehicle B. This provides an improved estimation of the vehicle's position uncertainty estimation needed for continuous positioning correction. We showed that a G-WhONet model does not fit a target vehicle as accurately as an S-WhONet trained for that vehicle. We also further demonstrated that the R-WhONet is able to improve the performance of the G-WhONet model to provide better uncertainty estimation accuracies closer to the S-WhONet model.

However, it is worth noting that one of the major reasons the R-WhONet model does not give estimations that exactly match that of the S-WhONet model is due to the limitation of the features of the new target vehicles used in the transfer learning training process. The first 50 seconds of the training data used for transfer learning might not have features which the model will encounter after adaptation. The continuous adaptation of the model might provide the opportunity to address this limitation. The WhONet model may be designed to be continuously triggered into recalibration mode for adaptation to the vehicle's new state This could be achieved through the  use of systems that are able to inform on the change in the states of the vehicle, using techniques that inform on the worn-out condition of the tyres or through information on the change in tyre pressures,. As the datasets used in this research do not provide information on exactly when the vehicular dynamics properties have changed, it has been challenging to implement such a system. However, this will be the subject of future research.

## 6. Appendix

***Table A1:*** *IO-VNB data subsets used in training G-WhONet*

| IO-VNB Dataset | Features |
|---|---|
| V-S1 | B-road (B4101), roundabout (x9), reverse (x5), hilly road, A4053 (ring-road), hard-brake, tyre pressure E |
| V-S2 | B-road (B4112, B4065), roundabout (x18), reverse drive (x8), motorway, dirt road, u-turn (x5), country road, successive left-right turns, hard-brake, A-roads (A4600), tyre pressure E |
| V-S3c | Roundabout(x4), A-road (A428), country roads, tyre pressure E |
| V-S4 | Roundabout (x14), u-turn, A-road, successive left-right turns, swift maneuvers, change in speed, night-time, A-road (A429, A45, A46), ring-road (A4053), tyre pressure E |
| V-St1 | Roundabout (x9), A-road (A452), B-road, car park navigation, tyre pressure E |
| V-M | Roundabout (x30), successive left-right turns, hard-brake (x21), swift maneuvers(x5), country roads, sharp turn left/right, daytime, u-turn (x1), u-turn reverse (x7), tyre pressure E |
| V-Y2 | Roundabout(x9), u-turn/reverse(x1), A-road, B-road, country road, tyre pressure E |
| V-Vta2 | Round About (×2), A Road (A511, A5121, A444), Country Road, Hard Brake, Tyre Pressure A |
| V-Vta8 | Town Roads (Build-up), A-Roads (A511), Tyre Pressure A |
| V-Vta9 | Hard-brakes, A–road (A50), tyre pressure A |
| V-Vta10 | Round About (×1), A—Road (A50), Tyre Pressure A |
| V-Vta13 | A-road (A515), country road, hard-brakes, tyre pressure A |
| V-Vta16 | Round-About (×3), Hilly Road, Country Road, A-Road (A515), Tyre Pressure A |
| V-Vta17 | Hilly Road, Hard-Brake, Stationary (No Motion), Tyre Pressure A |
| V-Vta20 | Hilly Road, Approximate Straight-line travel, Tyre Pressure A |
| V-Vta21 | Hilly Road, Tyre Pressure A |
| V-Vta22 | Hilly Road, Hard Brake, Tyre Pressure A |
| V-Vta27 | Gravel Road, Several Hilly Road, Potholes, Country Road, A-Road (A515), Tyre Pressure A |
| V-Vta28 | Country Road, Hard Brake, Valley, A-Road (A515) |
| V-Vta29 | Hard Brake, Country Road, Hilly Road, Windy Road, Dirt Road, Wet Road, Reverse (×2), Bumps, Rain, B-Road (B5053), Country Road, U-Turn (×3), Windy Road, Valley, Tyre Pressure A |
| V-Vta30 | Rain, Wet Road, U-Turn (×2), A-Road (A53, A515), Inner Town Driving, B-Road (B5053), Tyre Pressure A |
| V-Vtb1 | Valley, rain, Wet-Road, Country Road, U-T urn (×2), Hard-Brake, Swift-Manoeuvre, A—Road (A6, A6020, A623, A515), B-Road (B6405), Round About (×3), day Time, Tyre Pressure A |
| V-Vtb2 | Country Road, Wet Road, Dirt Road, Tyre Pressure A |
| V-Vtb5 | Dirt Road, Country Road, Gravel Road, Hard Brake, Wet Road, B Road (B6405, B6012, B5056), Inner Town Driving, A-Road, Motorway (M42, M1), Rush hour(Traffic) Round-About (×6), A-Road (A5, A42, A38, A615, A6), Tyre Pressure A |
| V-Vtb9 | Approximate straight-line motion, night-time, wet road, hard-brakes, A-road (A5), tyre pressure A |
| V-Vw4 | Round-About (×77), Swift-Manoeuvres, Hard-Brake, Inner City Driving, Reverse, A-Road, Motorway (M5, M40, M42), Country Road, Successive Left-Right Turns, Daytime, U-Turn (×3), Tyre Pressure D |
| V-Vw5 | Successive Left-Right Turns, Daytime, Sharp Turn Left/Right, Tyre Pressure D |
| V-Vw14b | Motorway (M42), Night-time, Tyre Pressure D |
| V-Vw14c | Motorway (M42), Round About (×2), A-Road (A446), Night-time, Hard Brake, Tyre Pressure D |
| V-Vfa01 | A-Road (A444), Round About (×1), B–Road (B4116) Day Time, Hard Brake, Tyre Pressure A |
| V-Vfa02 | B-Road (B4116), Round About (×5), A Road (A42, A641), Motorway (M1, M62) High Rise Buildings, Hard Brake, Tyre Pressure C |
| V-Vfb01a | City Centre Driving, Round-About (×1), Wet Road, Ring Road, Night, Tyre Pressure C |
| V-Vfb01b | Motorway (M606), Round-About (×1), City Roads Traffic, Wet Road, Changes in Acceleration in Short Periods of Time, Night, Tyre Pressure C |

***Table A2.*** *CUPAC data subsets used in training S-WhONet*

| **CUPAC Dataset** | **Features** | **Total Time Driven, Distance Covered, Velocity and Acceleration** |
|---|---|---|
| Alpha 1 | Inner-city, Parking lot, High traffic | 6.37 mins, 2.31 km, 0.518 to 39.35 km/hr, -0.25 to 0.16 g |
| Alpha 2 | Parking lot, Country road, Low | 8.33 mins, 4.46 km, 0.004 to 64.577 km/hr, -0.22 to 0.27 g |
| Alpha 3 | Inner-city, Country road, High traffic | 8.33 mins, 3.93 km, 0.0 to 55.174 km/hr, -0.21 to 0.21 g |
| Alpha 4 | Inner-city, High traffic | 8.33 mins, 3.99 km, 0.007 to 70.376 km/hr, -0.36 to 0.24 g |
| Bravo 1 | Inner-city, Medium traffic | 5.89 mins, 1.57 km, 0.0 to 40.374 km/hr, -0.27 to 0.16 g |
| Bravo 2 | Residential area, Road bumps, Low traffic | 16.92mins 7.46 km, 0.004 to 65.405 km/hr, -0.44 to 0.33 g |
| Bravo 3 | Residential area, Road bumps, Inner-city, Medium traffic | 25.38 mins, 12.51m, 0.0 to 69.48 km/hr, -0.68 to 0.31 g |
| Charlie 1 | Country road, Parking lot, Medium traffic | 4.86 mins,3.03 km, 0.004 to 62.68 km/hr, -0.26 to 0.21 g |
| Charlie 2 | Inner-city, Country road, Low traffic | 11.23 mins, 5.85 km, 0.004 to 58.104 km/hr, -0.38 to 0.27 g |
| Delta 1 | Highway, Residential area, Low traffic | 8.33 mins, 3.79 km, 0.004 to 53.654 km/hr, -0.19 to 0.25 g |

***Table A3.*** *CUPAC data subsets used for the longer GNSS outage scenario performance evaluation of the R-WhONet, S-WhONet, and S-WhONet*

| CUPAC Dataset | Total Time Driven, Distance Covered, Velocity and Acceleration |
|---|---|
| Charlie3 | 13.8 mins, 0.71 km, 0.004 to 77.252 km/hr, -0.42 to 0.25 g |
| Charlie 4 | 10.5 mins, 10.66 km, 0.007 to 58.694 km/hr, -0.34 to 0.31 g |
| Delta 2 | 59.9 mins, 96.5 km, 0.004 to 37.224 km/hr, -0.26 to 0.25 g |
| Delta 3 | 43.0 mins, 40.74 km, 0.004 to 60.75 km/hr, -0.36 to 0.25 g |